# Cascaded Face Alignment via Intimacy Definition Feature

Hailiang Li*, Kin-Man Lam*, Man-Yau Chiu, Kangheng Wu, Zhibin Lei

*Department of Electronic and Information Engineering, The Hong Kong Polytechnic University
Hong Kong Applied Science and Technology Research Institute Company Limited
Hong Kong, China
harley.li@connect.polyu.hk,{harleyli, edmondchiu, khwu, lei}@astri.org, enkmlam@polyu.edu.hk

***Abstract*—In this paper, we present a random-forest based fast cascaded regression model for face alignment, via a novel local feature. Our proposed local lightweight feature, namely intimacy definition feature (IDF), is more discriminative than landmark pose-indexed feature, more efficient than histogram of oriented gradients (HOG) feature and scale-invariant feature transform (SIFT) feature, and more compact than the local binary feature (LBF). Experimental results show that our approach achieves state-of-the-art performance when tested on the most challenging datasets. Compared with an LBF-based algorithm, our method can achieve about two times the speed-up and more than 20% improvement, in terms of alignment accuracy measurement, and save an order of magnitude of memory requirement.**

***Index Terms*—Cascaded face alignment, random forest, intimacy definition feature.**

## I.INTRODUCTION

FACE alignment is an active research topic in computer vision. It is often used as an early, but crucial, step to other important tasks for face analysis, such as emotion and expression recognition [9], face recognition [10], and face hallucination [11]. It is also used in many other applications, such as human-machine interactions, video conferencing, gaming and animation, etc., and has received increasing attention from the computer-vision research community. Face alignment is a process to locate facial key-points or landmarks (e.g. the eyebrows, eye corners, and mouth corners, see Fig. 1) from a given face image.

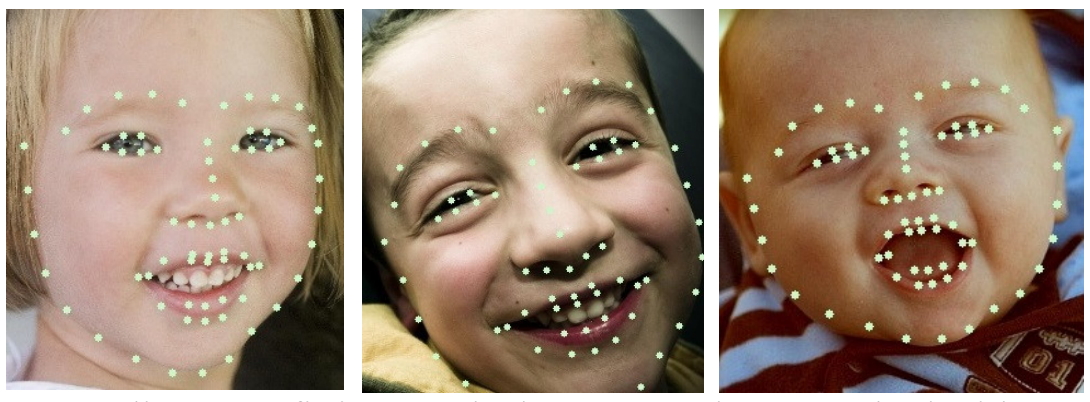

Fig. 1. Face alignment fitting results by proposed IDF method with 68 points (the Helen dataset [21]).

A majority of face-alignment methods assume that the face-bounding box, for a face image, is known at both the training and the fitting stages. The face-bounding box is usually obtained either through a face-detection algorithm, such as the Viola-Jones [12] face detector, or from manual annotations, i.e. the ground truth. The set of coordinates of the facial landmarks is usually referred to as the face shape. Various machine-learning algorithms have been proposed to estimate the face shape. Traditional methods for face alignment usually involve the classic pioneering works, including active shape model (ASM) [3] and active appearance model (AAM) [4]. Both ASM and AAM are statistical models. ASM represents the shape of an object, while AAM represents both texture and shape. Constrained local models (CLM) [24, 25, 27, 28, 31] attempt to model shape prior, similar to AAM, by assuming that face local appearance information and global face-shape patterns lie in a linear subspace spanned by bases, learned from principal component analysis (PCA). In [26], face-shape fitting is formulated as a non-linear minimization problem, which sets a target of minimizing the error (i.e. the average distance of all the respective landmarks normalized by the inter-pupil distance) between the model instance and the model of a given image, with respect to the model parameters that control the shape and appearance variations of faces. In [26], an extension to the inverse compositional image-alignment algorithm [29] was proposed, which decouples shape from appearance. The method forms a computationally efficient AAM framework. Usually, a CLM model is composed of three main parts: a point distribution model (PDM), patch experts which perform matching for local patches around landmarks of interest, and a final fitting process. Different fitting strategies have been used in CLM. Regularized landmark mean shift (RLMS) [28] is a popular strategy, which estimates the rigid and non-rigid parameters by minimizing the misalignment error of all the landmarks, regularized by overly complex or unlikely shapes. In [27], a local neural filed (LNF) patch expert was proposed, which learns the similarity and long-distance sparsity constraints to derive relationships between neighboring pixels and longer-distance pixels. The method achieves state-of-the-art performance, compared to the traditionally methods based on CLM. Similar work in [32], the authors proposed an exemplar-based graph matching (EGM) framework for face alignment, in which the response mappings of all the facial landmarks are fitted by selecting from a pool of training exemplar poses.

However, these models have limited expressive power to capture all possible complex and subtle face variations, due to variations in expression, illumination, pose, etc. Furthermore, due to the computational requirements for the inverse of Hessian matrix and the Jacobian matrix [6], it is very hard for those algorithms using CLM to improve their speeds exponentially. Only training-based regressions can achieve the efficiency required for real-world applications, such as smart mobile phones.

In the past few years, a new family of face-alignment algorithms has emerged [1, 5, 6, 8, 13], which directly learns regressors from image feature descriptors to the target shape. These regression-based methods are gaining popularity, due to their excellent performance and high efficiency in the face-alignment task. Pose-indexed feature [6, 8, 13], whose index provides some clue about the hierarchical structure of the shape, is employed to enhance efficiency. In [6], the handcrafted scale-invariant feature transform (SIFT) feature is used for accurate fitting. As inspired by the works in [1, 6], an efficient and discriminative feature is a crucial element for random-forest-based cascaded face-alignment algorithms. In this paper, we propose a novel and efficient feature, which can be incorporated into other face-alignment frameworks, to further boost their performance.

Recently, deep learning based models have been emerging as hot research topics and successfully applied to many computer vision tasks such as generic object detection and classification [33, 34, 35], handwritten digit recognition [38], RGB-D object recognition [39], image super-resolution [41, 42, 43], visual tracking [44], attribute prediction [45], face alignment [36, 37, 40] and so on.

In [40], the authors try to improve face landmark detection through multi-task learning and design a tasks-constrained deep model with task-wise early stopping to increase the learning convergence rate. Authors in [37] exploit using deep network to learn feature-to-pose mapping functions by combining the cascaded framework of regressing on pose-indexed features and deep learning. To better initialize facial poses, in [36], the authors present a global exemplar-based deep auto-encoder network (GEDAN) to increase the capacity of pose estimation for handling large pose variation by incorporating several exemplars at the top layer which builds a non-linear regression model. Although these brute-force-style deep learning approaches has achieved promising performance in terms of fitting accuracy, but their heavy computation is a big obstacle to real-world applications where there is less chance with graphics processing unit (GPU) speed-up.

After an introduction to face alignment, the remainder of this paper is organized as follows. In Section II, we will review the random-forest classifier, and the random-forest-based cascaded face-alignment approach. In Section III, a feature derived from the shape-index feature, named intimacy definition feature (IDF), will be presented. Then, our proposed IDF-based cascaded random-forest face-alignment algorithm will be described and analyzed. Section IV will evaluate our proposed method and compare it with recent fast local binary feature (LBF)-based method. Section V will discuss how to cluster the training samples into subspaces for selecting representative shapes to form initialization samples. Experiment results and parameter settings will be presented in Section VI, and conclusions and future work are given in Section VII.

## II. Random Forests for Landmark Localization

The landmark localization algorithm is important for face recognition and other related applications, which require extraction of local features at some specified feature points or landmarks in a face. For face alignment, numbers of points, usually between 17 and 68, are selected and searched from a face image. An example of the landmarks is shown in Fig. 1, which has 68 points located around the eyes, nose, lips, and face contour. These feature points, which are useful for discriminative and generative analysis, carry the most significant information about a face. Based on these feature points, a model can then be learned from numbers of landmark-labeled face images, used for facial-shape estimation for unseen face images.

Recently, there have been three main approaches for face alignment, methods based on the active appearance models (AAM) [4] that build parametric models of appearance, deep learning based models [36, 37, 40] and regression-based models that directly model a mapping from appearance to shape [1, 5, 6, 8, 13]. In our algorithm, we adopt the regression-based approach, based on the cascaded shape-regression framework firstly proposed by Dollar et al. in [8]. Different from other methods, this approach progressively refines the initial shape estimate in several stages directly from appearance, without requiring to learn any parametric shape or appearance models. As the cascaded regression algorithms are mainly based on random forests, we give a brief review of the main principles of random forest and cascaded-shape regression in this section.

### II.1 Random Forest

In recent years, random forests [14] (RFs) have emerged as very useful classifiers for a large variety of computer-vision tasks, including object detection [16], data clustering [17], image super-resolution [18, 19], etc. This method is relatively simple, and has many merits that make it particularly interesting for computer-vision problems: (i) efficiency in both training and prediction, (ii) inherent unsupervised classification capability for multi-class problems, (iii) suitability for parallel processing, and (iv) good performance for high-dimensional data.

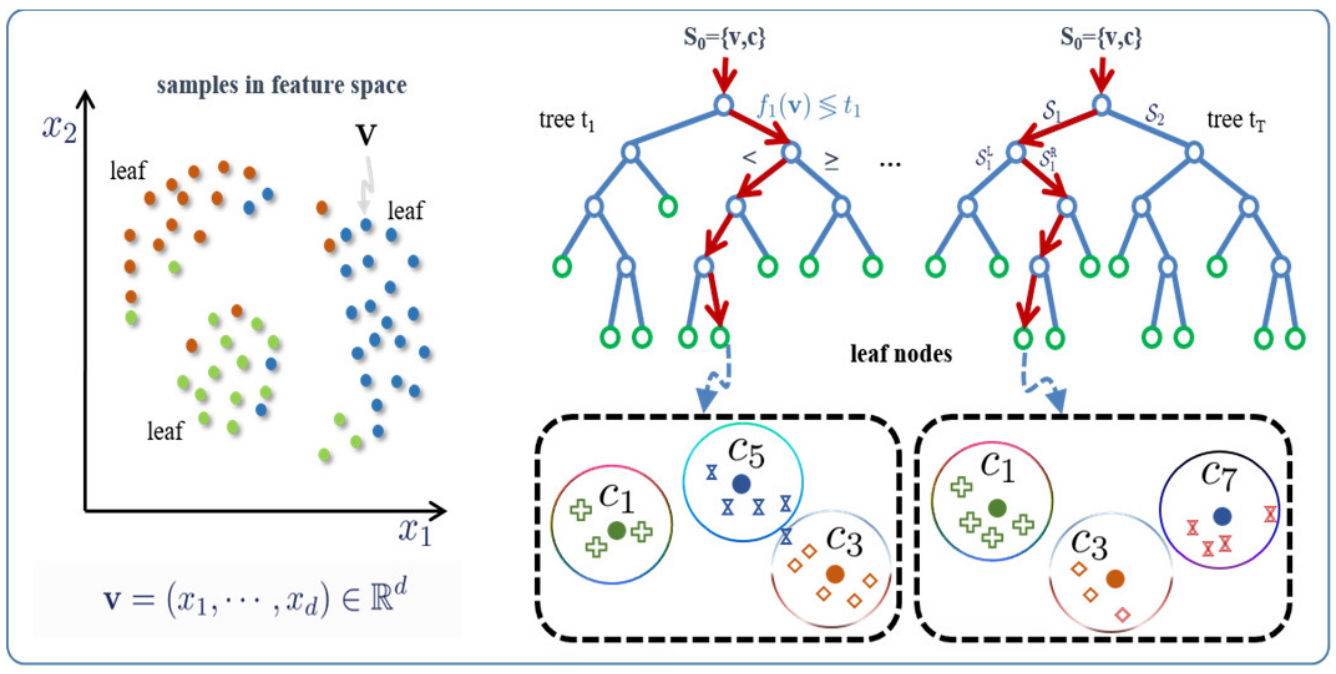

Fig. 2: Random forest for clustering data.

A random forest is an ensemble of $T$ binary decision trees $T^t(x): V \to R^K$, where $t$ is the index of the trees, $V \in R^M$ is the $M$-d feature space, and $R^K$=[0, 1]$^K$ represents the space of class probability distributions over the label space $Y = \{1, . . . , K\}$, as shown in Fig. 2.

In the testing/prediction stage, each decision tree returns a class probability $p_t(y|\boldsymbol{v})$ for a given test sample $\boldsymbol{v} \in R^M$, and the final class label $y^*$ is then obtained via averaging:

$$y^* = \arg\max_y \frac{1}{T}\sum_{t=1}^{T} p_t(y|\boldsymbol{v}). \quad (1)$$

A splitting function $s(v; \Theta)$ is typically parameterized by two values: (i) a feature dimension $\Theta_1 \in \{1, . . . , M\}$, and (ii) a threshold $\Theta_2 \in R$. The splitting function is defined as follows:

$$s(\boldsymbol{v}; \Theta) = \begin{cases} 0, & \text{if } \boldsymbol{v}(\Theta_1) < \Theta_2, \\ 1, & \text{otherwise}, \end{cases} \quad (2)$$

where the outcome defines to which child node the sample $\boldsymbol{v}$ is routed, and 0 and 1 are the two labels belonging to the left and right child nodes, respectively. Each node chooses the best splitting function $\Theta^*$ out of a randomly sampled set $\{\Theta^i\}$ by optimizing the following function:

$$I = \frac{|L|}{|L|+|R|} H(L) + \frac{|L|}{|L|+|R|} H(R), \quad (3)$$

where $L$ and $R$ are the sets of samples that are routed to the left and the right child nodes, respectively, and $|S|$ represents the number of samples in the set $S$. During the training of a random forest (RF), each decision tree is provided with a random subset of the training data (i.e. bagging), and is trained independently of other trees. Training a decision tree involves recursively splitting each node, such that the training data in the newly created child nodes are clustered according to class labels. Each tree is grown until a stopping criterion is reached (e.g. the number of samples in a node is less than a threshold or the tree depth reaches a maximum value), and the class probability distributions are estimated in the leaf nodes. $H(S)$ is the local score for a set of samples ($S$ is either $L$ or $R$), which normally is calculated using entropy as in (4), but it can be replaced by variance [1] or the Gini index [14].

$$H(S) = -\sum_{k=1}^{K}[p(k|S) * \log(p(k|S))] \quad (4)$$

where $K$ is the number of classes, and $p(k|S)$ is the probability for class $k$, which is estimated from the clustered set $S$.

### II.2 Cascaded shape regression

Many face alignment methods work under a cascaded framework, where an ensemble of $N$ regressors operates in a stage-by-stage manner, which are referred to as stage regressors. This approach was first explored in [8]. At the testing stage, the input to a regressor ($R_t$) at stage $t$ is a tuple $(I, S^{t-1})$, where $I$ is an image and $S^{t-1}$ is the shape estimate from the previous stage (the initial shape $S^0$ is typically the mean shape of the training set). The regressor extracts features with respect to the current shape estimate, and regresses a vector of shape increment as follows:

$$S_t = S_{t-1} + R_t(\phi_t(I, S_{t-1})), \quad (5)$$

where $\phi_t(I, S_{t-1})$ is referred to the feature extraction function, such as the pose-indexed features, i.e. they depend on the current shape estimate. The cascade progressively infers the shape in a coarse-to-fine manner – the early regressors handle large variations in shape, while the later ones ensure small refinements. After each stage, the shape estimate resembles the true shape closer.

In our algorithm, the feature mapping function $\phi_t(I, S_{t-1})$ generates the local IDF values derived from the pose-indexed feature. There is an assumption, proved by intensive experimental results, that the shape increments have close correlation with the local features of the landmarks, which define the face shape. Thus, given the features and the target shape increments $\{\Delta S_t = S - S_{t-1}\}$, we can learn the linear projection matrix $R_t$. Most cascaded regression models [1, 5, 6, 8, 13] have a similar workflow, as shown in Fig. 5.

## III. Intimacy Definition Feature Based Cascaded Regression Model

In this section, we will first introduce a new feature, which is efficient for local pattern representation and matching, based on measuring the degree of intimacy (DoI) between two features.

### III.1 Efficient Metric on Intimacy Definition Feature

To explain the features, we replace a feature with a member in a family tree, and we measure the DoI between two family members. The relationships between the family members are represented by using a binary family tree, which is similar to one tree in random forests scheme, as shown in Fig. 3. The DoI between two family members can be computed by their respective intimacy definition feature (IDF) values. In Fig. 3, the DoI between David and Daniel should be stronger than that between David and Denis. This is because David and Daniel have the same father, while David and Denis do not have the same father but they share the same grandfather only. The way to let the computer learn the DoI value, between any two members in the same generation or level in the hierarchical family tree, is to digitize the DoI values. This means to set values to all the nodes in the family tree and define a distance metric between any two of the members.

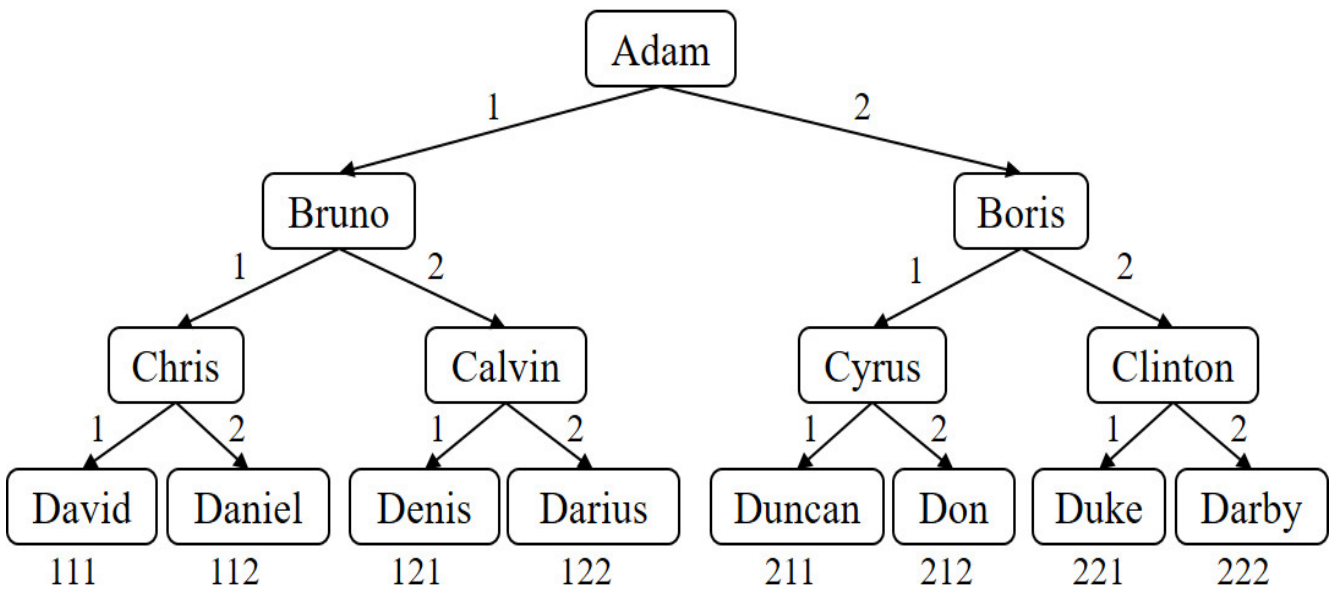

Fig. 3: A family tree shows the degree of intimacy between two persons in the 4th generation.

As we can see in the family tree in Fig. 3, two persons, who share more recent parents, should be more intimate than those who share relatively distant parents, as described in the previous example. However, how can a computer know this intimacy, based on this logic comparison operation? In this paper, we propose a simple, yet efficient, method to compare the DoI values between two members in the same generation. We firstly set two persons in the same generation with values of very small difference, for example, we set 1 and 2 as the respective *path values* to the two offspring nodes (e.g. David is the younger brother so his *path_value* is 1, while Daniel is the older brother so his *path_value* is 2) in the full binary family tree. Then, we set a relatively larger value, e.g. 10, to the *generation value k* for each generation level. Each node (except the root node) can then be encoded by summing up all the corresponding level weights along the path from the root to the node of a member concerned, where a level weight of a node is computed by multiplying the value of the node and its corresponding generation value *k*. We name this as the intimacy definition feature (IDF) value of the node or family member, which can be calculated as follows:

$$IDF = \sum_{l=1}^{L} path_value_l * k^l, \tag{6}$$

where $L$ is the total number of levels in the family tree. Therefore, the IDF value of David can be encoded as: 111 ($1\times10^2+1\times10^1+1\times10^0$), and Daniel with IDF value: 112 ($1\times10^2+1\times10^1+2\times10^0$). We can also encode Denis as IDF value: 121 ($1\times10^2+2\times10^1+1\times10^0$). The intimacy distance between David and Daniel is 1 (1 = abs(111−112)), and the distance between David and Denis is 10 (10 = abs(111−121)). The distances show that the intimacy between David and Daniel should be greater than that between David and Denis. Based on the proposed IDF, we can compute the DoI value between the IDF values of two family members in the family tree. Through the family tree as constructed in Fig. 3, the family members (nodes) can be replaced by visual features, which are encoded by IDF values. Consequently, the similarity between two family members (nodes) can be computed by measuring their DoI.

In our study, we found that this simple, yet efficient, feature can be computed by traveling a tree in random forests, which can achieve promising performance, in terms of both accuracy and speed, as shown in Section IV. When using the encoded feature values for linear regression on the leaf nodes for prediction, for more reliable and better performance, the feature is normalized as follows:

$$nomalized_IDF = \frac{(IDF - IDF_{min})}{(IDF_{max} - IDF_{min})}, \tag{7}$$

where $IDF_{min}$ and $IDF_{max}$ are the minimum and maximum IDF values, respectively, in the same level under consideration. Using our example, the range of the IDF values in the binary tree is [100, 222], i.e., $IDF_{min}$ = 100 and $IDF_{max}$ = 222. Therefore, based on (7), the normalized IDF value for David (111) can be calculated as: (111−100)/(222−100) = 0.090164.

### III.2 Derive IDF feature from pose-indexed feature

For each stage, the whole feature vector $\Phi_t$ is a concatenation of a set of independent local-feature mapping functions: $\phi_t(I, S_{t-1})$, which is the local IDF values, derived from the pose-indexed feature in our algorithm. A pose-indexed feature is the value of two pixels' intensity difference. For every landmark point, those two pixels used to compute the pose-indexed value are chosen with two randomness in the random forest splitting rule, which means that they are randomly sampled from a number of candidate pixels (e.g. 500) and the threshold is also randomly selected. The positions of the pixel pair and the threshold to be used are decided, based on maximizing the information gain obtained when splitting all the samples in a node into its left and right nodes.

Same as algorithm LBF [1], we discard such learned local pose-indexed features since they are not sufficiently discriminative, or do not encode the path of a sample along a tree explicitly. Instead, we encode the path of a sample along a tree ended at a leaf node, using our proposed IDF values. As described in in Fig. 3, each IDF value encoded in a leaf-node is one float scalar, which can achieve highly dimensionality reduction compare to the sparse but high dimensional LBF features generated from the proposed method in [1].

All the IDF features are concatenated to form a global feature mapping function $\Phi_t$ for learning a global linear projection, i.e. the regressor $R_t$, in the next step.

All the pixel pairs are sampled from the neighborhood area which is centered at each landmark point. Different faces should normally be aligned with different rectangles, but the same neighborhood size can be used for the landmark points of different faces. The idea of our pose-indexed feature is described in Fig. 4.

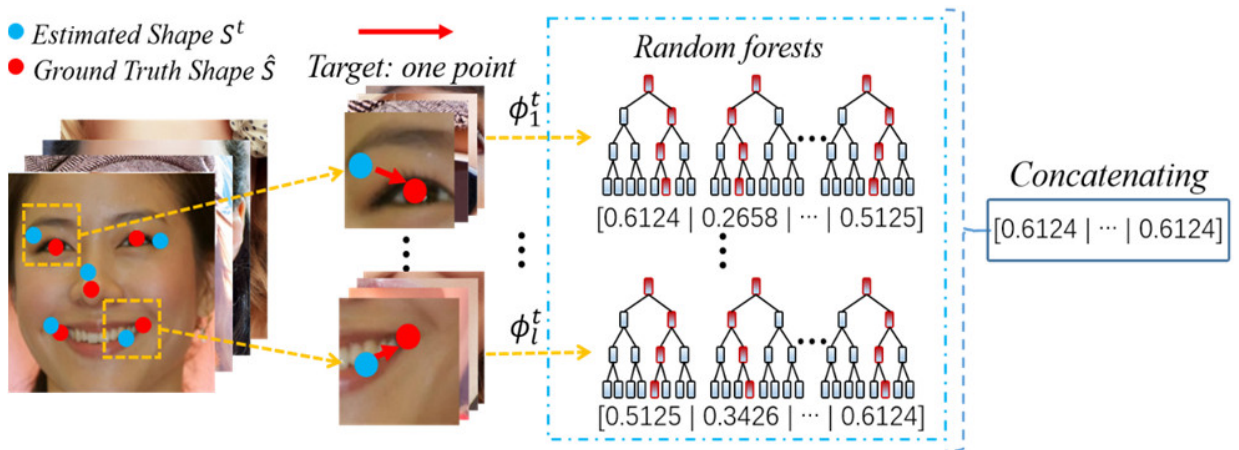

Fig. 4: The process of IDF-based feature vector extraction

In both training and prediction stages, the neighborhood size for each landmark can be reduced, when moving from one cascade to another cascade. Therefore, the cascaded shape regression operates from coarse to fine progressively.

### III.3 Cascaded shape regression based on IDF feature

A similar work to our proposed algorithm is the LBF-based method in [1], which improves the supervised descent method (SDM) [6] used in linear regression. In [1], random forests were used for training, so as to minimize the alignment error for the respective landmarks with LBF, rather than the pose-indexed feature in the leaf nodes. LBF is a local feature, which is coded as a binary array, by placing the value '1' for leaf nodes, where samples fall into them eventually while traversing a tree in random forests, and the value '0' otherwise. Each landmark is coded individually, and the local features are concatenated to form a global feature vector, which is then learned by using ridge regression (i.e., linear regression with $L^2$ regularization). Rather than using LBF, our proposed IDF is employed in the cascaded alignment framework, as depicted in Fig. 5. To further improve the performance, we refine the shape initialization by using the *k*-means clustering algorithm. The success of LBF [1] method is due to its feature-learning step, where features are explicitly learned for the given specific task. Due to the sparse nature of the feature vector of LBF, the testing phase can be reduced to traversing the forest, and performing simple table look-ups and additions. The authors in [1] claimed that LBF method can achieve an impressive speed of 3,000 fps, the fastest approach.

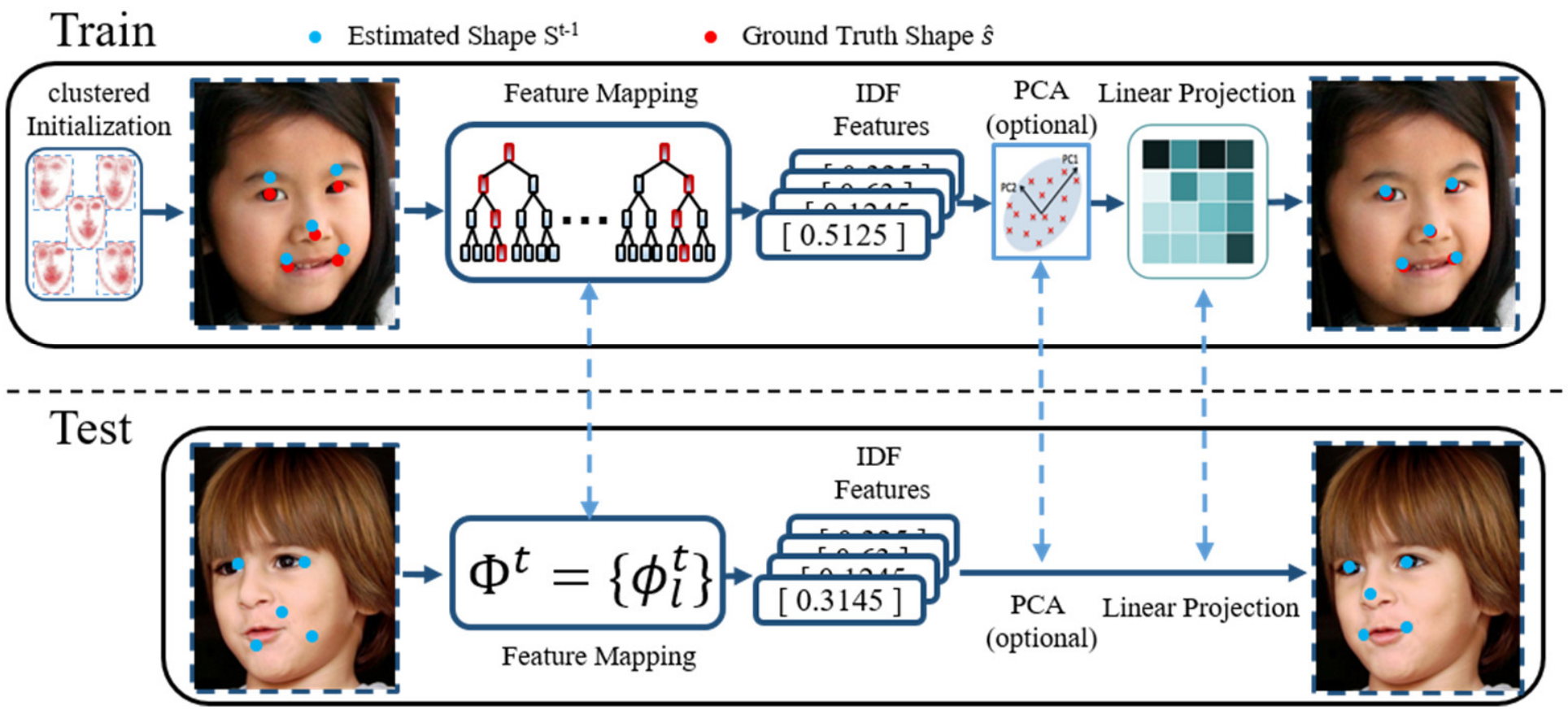

Fig. 5: An overview of the workflow for IDF-based cascaded regression face alignment.

However, LBF has a high dimensionality. Assume that the number of landmarks (or forests) for a face is $l$, the number of trees for a forest is $t$, and the depth of a tree is $d$. The dimensionality of LBF will then be $l \cdot t \cdot 2^{(d-1)}$. For a normal setting of $l = 68$, $t = 10$, and $d = 7$, the feature dimension is $68 \times 10 \times 2^{(7-1)} = 43{,}520$, which is relatively high. Usually, with more and deeper trees, the alignment errors will become smaller. However, the high dimensionality of LBF restricts it from using deeper trees. Although the feature is sparse, its high dimensionality imposes a high burden on the computation of linear regression and the storage requirement. An intuitive way to solve the problem is to employ PCA to reduce the dimensionality. However, LBF is a binary, sparse feature, and carries labelling information, which makes PCA not acceptable for the feature. To avoid the computational complexity, the LBF-based approach has to limit the tree depth to relative small value, e.g.: 5, which means that there are, at most, 16 leaf nodes in each tree. Consequently, this heavily restricts its capability on classification and prediction.

Compared to the pixel pose-indexed feature [13], LBF is more discriminative because it explicitly encodes the full path, from the root to the leaf node of each sample. Although LBF is discriminative, it is hard to greatly improve its performance because of its high dimensionality when using deeper trees. To improve the performance, an intuitive way is to replace LBF with another more compact and efficient *index feature*, which encodes the path of a sample along a tree. However, the performance is very poor, because index values are similar to labels, which inclines more to results with over-fitting. A simple analysis in Fig. 3 can help describe the problem of using an *index feature*. Suppose that we simply set the indices for David, Daniel, and Denis at 1, 2, and 3, respectively, as shown in Fig. 3. With these values, we can find that the DoI value between David and Daniel is the same as that between Daniel and Denis. However, from Fig. 3, we know the intimacy between David and Daniel should be closer than between Daniel and Denis.

Our algorithm is based on extracting the IDF value at each facial landmark, using the full binary family tree. With the IDF values, leaf nodes can be compared based on their DoI. The main contribution of this paper is that the efficient IDF feature is proposed to replace LPF. This can greatly reduce the feature dimensionality, while a promising performance can still be achieved. More importantly, our algorithm runs much faster and requires less memory than that using LPF. For example, for $l = 68$, $t = 10$, and $d = 7$, the feature dimensionality of IDF is $68 \times 10 \times 1 = 680$, rather than 43,520 ($= 68 \times 10 \times 64$) for LBF. In other words, the dimensionality is reduced by 64 times.

## IV. Validation Results and Comparison to LBF Method

To validate the effectiveness, efficiency, and less memory usage of our proposed IDF method on face alignment compare to LBF [1] method, we did intensity experiments on the public datasets.

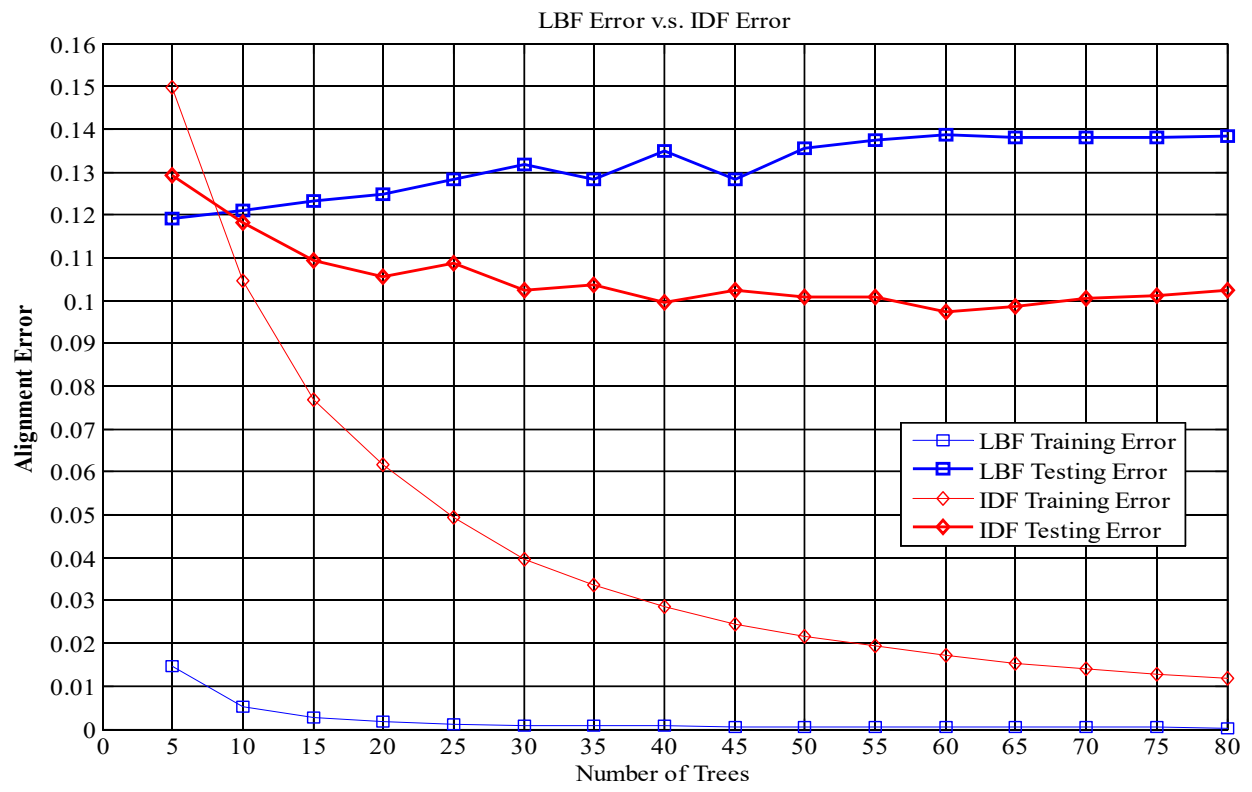

Fig. 6: A comparison of the alignment errors of the IDF and LBF algorithms on the LFPW dataset [20], with tree depth = 7,training: 1000, testing: 300.

To demonstrate the effectiveness of IDF for face alignment, we set tree depth, maximum number of stages, and number of landmarks at 7, 7, and 68, respectively, and measure the respective alignment errors using LBF and IDF. Fig. 6 shows the alignment errors in the training and testing stages, based on the LFPW dataset [20], with different numbers of trees. From the results, we can see that our proposed IDF algorithm can achieve, on average, an error of around 0.10, when the number of trees is more than 10, while the minimum error achieved by the LBF-based algorithm is 0.12. Therefore, our algorithm can achieve an improvement of about 20%, in terms of alignment error, when compared to the LPF-based algorithm.

| | **Number of Trees** | | | | | | | | | |
|---|---|---|---|---|---|---|---|---|---|---|
| stage | 5 | 10 | 20 | 30 | 40 | 50 | 60 | 70 | 80 | *Avg.* |
| 1 | 0.1765 | 0.1714 | 0.1630 | 0.1583 | 0.1583 | 0.1583 | 0.1533 | 0.1495 | 0.1485 | **0.1597** |
| 2 | 0.1411 | 0.1341 | 0.1300 | 0.1315 | 0.1315 | 0.1315 | 0.1397 | 0.1387 | 0.1390 | **0.1352** |
| 3 | 0.1276 | 0.1251 | 0.1252 | 0.1292 | 0.1292 | 0.1292 | 0.1390 | 0.1382 | 0.1386 | **0.1312** |
| 4 | 0.1232 | 0.1226 | 0.1240 | 0.1287 | 0.1287 | 0.1287 | 0.1389 | 0.1381 | 0.1385 | **0.1301** |
| 5 | 0.1209 | 0.1217 | 0.1235 | 0.1285 | 0.1285 | 0.1285 | 0.1388 | 0.1380 | 0.1384 | **0.1296** |
| 6 | 0.1198 | 0.1212 | 0.1234 | 0.1284 | 0.1284 | 0.1284 | 0.1388 | 0.1380 | 0.1384 | **0.1294** |
| 7 | 0.1193 | 0.1209 | 0.1233 | 0.1283 | 0.1283 | 0.1283 | 0.1388 | 0.1380 | 0.1384 | **0.1293** |

Table-1: Alignment errors at different stages, with different number of trees, based on the LBF algorithm.

| | **Number of Trees** | | | | | | | | | |
|---|---|---|---|---|---|---|---|---|---|---|
| stage | 5 | 10 | 20 | 30 | 40 | 50 | 60 | 70 | 80 | *Avg.* |
| 1 | 0.1924 | 0.1915 | 0.1873 | 0.1937 | 0.1886 | 0.1914 | 0.1826 | 0.1810 | 0.1856 | **0.1882** |
| 2 | 0.1636 | 0.1583 | 0.1472 | 0.1462 | 0.1412 | 0.1360 | 0.1312 | 0.1318 | 0.1326 | **0.1431** |
| 3 | 0.1540 | 0.1412 | 0.1294 | 0.1283 | 0.1206 | 0.66 | 0.1112 | 0.1129 | 0.1136 | **0.1254** |
| 4 | 0.1445 | 0.1309 | 0.1188 | 0.1192 | 0.1119 | 0.1091 | 0.1041 | 0.1059 | 0.1073 | **0.1168** |
| 5 | 0.1380 | 0.1249 | 0.1136 | 0.1142 | 0.1076 | 0.1057 | 0.1010 | 0.1032 | 0.1049 | **0.1126** |
| 6 | 0.1334 | 0.1200 | 0.1114 | 0.1104 | 0.1051 | 0.1039 | 0.0990 | 0.1015 | 0.1034 | **0.1098** |
| 7 | 0.1291 | 0.1180 | 0.1093 | 0.1089 | 0.1036 | 0.1024 | 0.0974 | 0.1005 | 0.1025 | **0.1080** |

Table-2: Alignment errors at different stages, with different number of trees, based on the IDF algorithm.

Another factor we should consider is the number of trees required to achieve a specific alignment error. From Fig. 6, we can see that using about 10 trees for our algorithm can achieve even smaller errors than that of LBF using more than 80 trees. As shown in Table 1 and Table 2, although LBF performs better in the first 3 stages, IDF can always achieve better performance at later stages, since its alignment error converges steeper than LBF. In other words, IDF converges faster in the coarse-to-fine search of the true landmarks, with a higher discriminative power.

Fig. 7(a) shows the alignment errors of the LBF and IDF methods, with different numbers of stages used. We can see that the curve for IDF is much steeper than that for LBF, which means that the IDF feature converges to lower errors faster, and is more discriminative than LBF at later stages. An explanation for this is that the IDF value is represented as floating point numbers, and has a stronger representation than LBF represented by binary numbers. Fig. 7(b) shows the alignment errors of IDF with a larger number of stages used. To make a balance between computational complexity and fitting accuracy, using 7 stages is a compromise. Therefore, in the rest of this paper, our algorithm uses 7 stages of cascade.

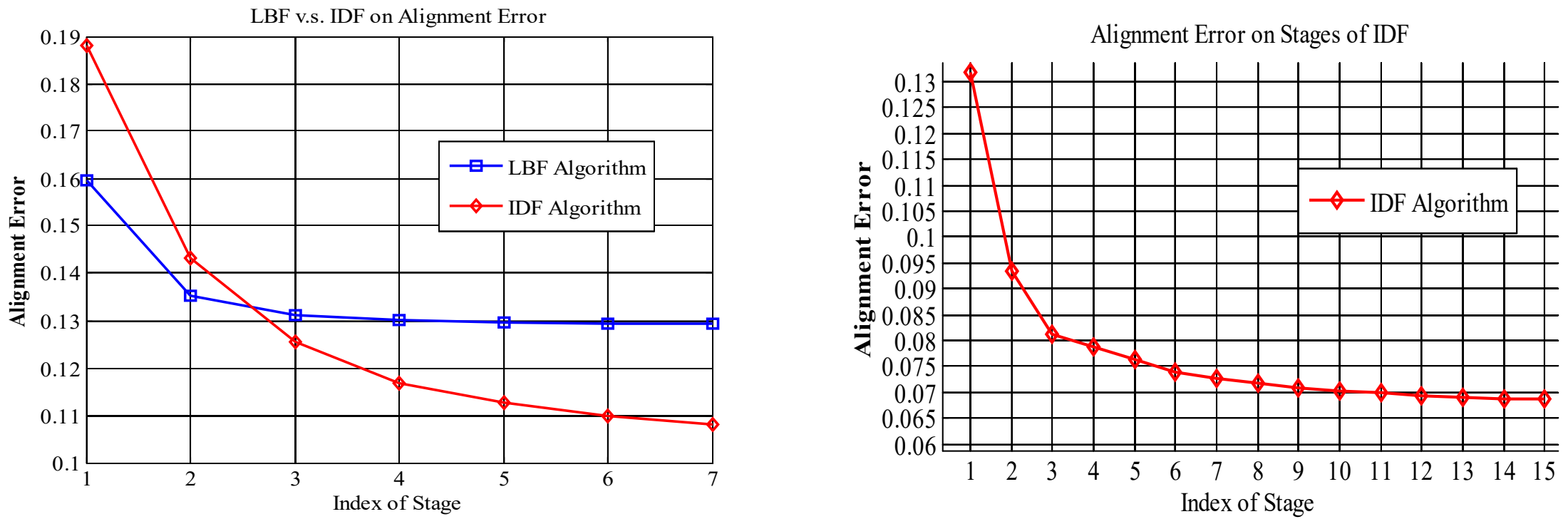

Fig. 7: Alignment errors with different numbers of stages of cascade: (left) based on LBF and IDF, up to 7 stages, and (right) based on IDF only, up to 15 stages (tree depth = 7, LFPW dataset [20]).

Having analyzed the LBF algorithm, we found that, in the prediction stage, feature extraction and linear regression take up about 30% and 70% of the total computation, respectively. Since our proposed IDF is derived from the pose-indexed feature, which takes up less computation in the extraction. Furthermore, IDF has its dimensionality an order of magnitude lower than that of LBF, so the computational complexity for linear regression (the *LibLinear* package is used for IDF and LBF) is greatly reduced. As shown in Fig. 8, the number of frames processed per second, based on IDF, is about 2 times faster than LBF, with the same setting.

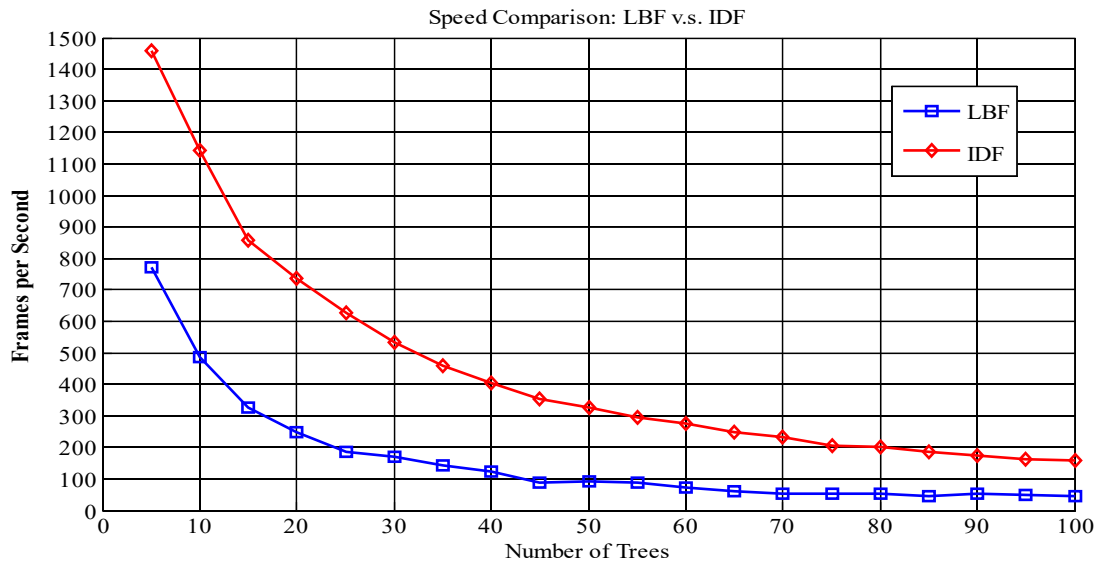

Fig. 8: The speeds in terms of number of frames per second for the IDF and LBF algorithms (tree depth = 7, Helen dataset ).

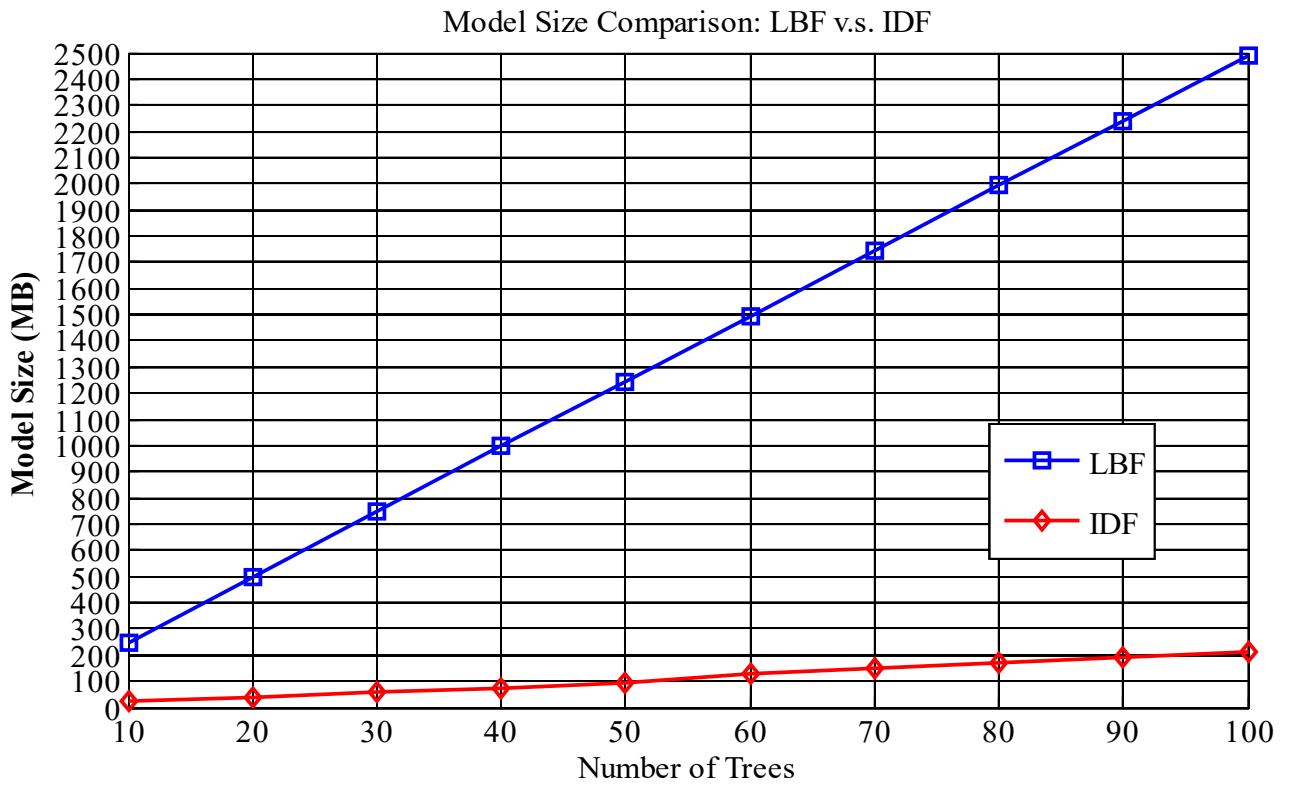

Fig. 9: Memory requirements (MB) of IDF and LBF with different numbers of trees (tree depth: 7, Helen dataset).

When the tree depth increases, the feature dimensionality of LBF increases exponentially, while the IDF algorithm increases linearly. In addition to computational complexity, memory requirement is also an important issue for real applications. Because of the high dimensionality of LBF, more weights for linear regression need to be saved for the regression model.

That means the LBF-based method needs much more memory in the prediction step, as all the regression models for the cascaded stages are to be kept in memory, as shown in Fig. 9.

## V. Training with Representative Shapes from Clustered Subspaces

A limitation of the cascaded regression model is that the approach is sensitive to the initial shape, which means that using a common mean shape, it may not be possible to obtain good performance on some images with unseen face profiles. In [5], a conditional regression forest was proposed for face alignment, in which annotated samples are used to train a classifier to detect the face pose with discriminative features inside and outside the face-bounding boxes. Based on the annotated face poses, a number of cascade regression forest models are trained, instead of a single model only. In the testing stage, once the face pose has been detected using the pose detector, the probability of the head pose is estimated from the testing image, and the corresponding trees are selected for later face alignment.

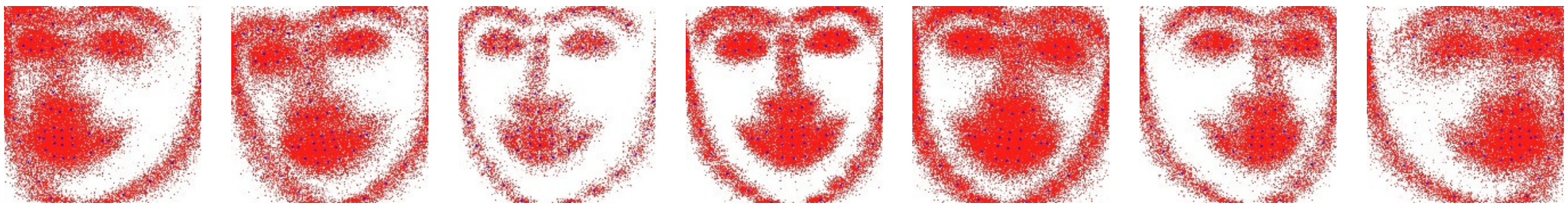

Fig. 10: Results from clustering with 7 groups of faces with different poses.

In [2], a pose detector is employed for estimating initial shapes, based on the *k* nearest neighbors selected from training samples, which uses two efficient and effective features, namely the histogram of oriented gradients (HOG) [22] and local binary patterns (LBP) [23], for searching example face images with a similar pose and texture appearance to the query face, respectively. The local appearance of feature points can be accurately approximated with locality constraints. Therefore, with the searched training faces, which have similar poses and textures to a testing face, more accurate initial shape model can be constructed in the prediction stage.

Different from the two above-mentioned methods [5, 6], our algorithm does not use any pose detector that considers similar texture. The training samples are selected, based on the similar shapes spanned in face-shape subspaces. As using random initial shapes in the training phase can improve the generalization capability of the alignment method, this means that the trajectory of face alignment, during the regression stages in the prediction process, can be learnt from training samples. Intuitively, for a face with a large pose, the shape trajectory of a left-pose face cannot be learnt from a right-pose face. Therefore, training samples selected for the initial model should have similar poses, which can help to learn the pose information implicitly. In [5], a face dataset with different poses and with 10 landmark points was created. However, we consider 68 landmark points in face images, and we will evaluate our algorithm using some standard public datasets, such as the LFPW dataset [20] (811 training + 224 testing images taken under unconstrained conditions (in the wild) with large variations in the pose, expression, illumination and with partial occlusions) and Helen dataset [21] (2000 training + 330 testing, the images exhibit a large variety in appearance, such as pose, expression, ethnicity, age and gender, as well as the general imaging and environmental conditions). The dataset can be labelled manually, as it was in [5], so that the learning will be more precise. In this paper, we have devised a more efficient scheme for this training process. This is important because the tedium of labeling faces and their poses is challenging and the work may be imprecise. For example, it is difficult to decide on a face with a pose (e.g. at 45-degree angle) from two clusters (e.g. 30-degree angle and 60-degree angle). It is often an ambiguous task for human eyes. In [2], although *k* nearest neighbors similar to the test face are searched with locality constraints, a relatively narrow subspace may be produced, based on the *k* training samples. This may spoil the generalization capability of the learned model.

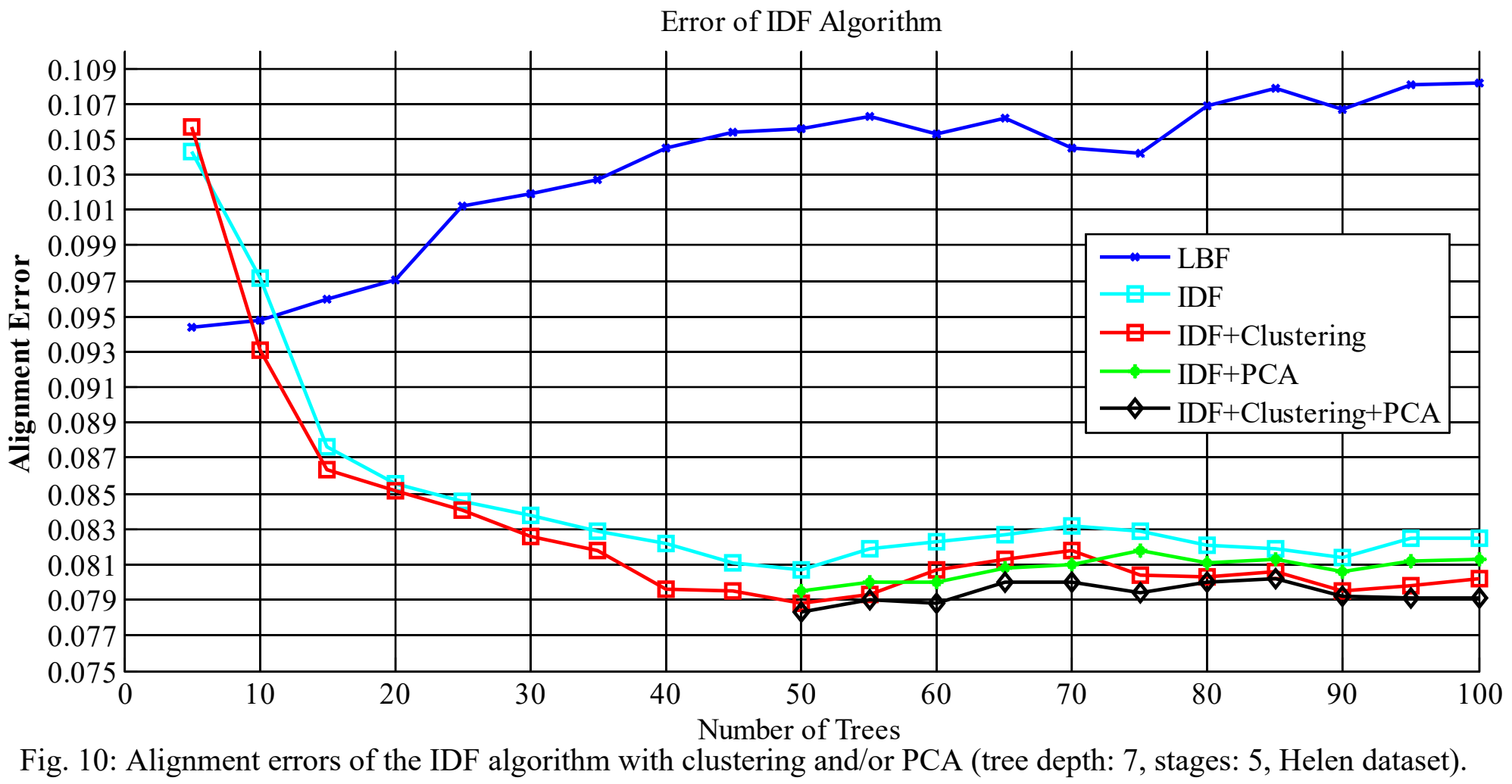

Fig. 10: Alignment errors of the IDF algorithm with clustering and/or PCA (tree depth: 7, stages: 5, Helen dataset).

We use the $k$-means algorithm to group the training samples into a number of clusters, as shown in Fig. 10. Then, for each test face image, instead of learning the initial shape using the whole training dataset, we select example shapes only from the cluster with a similar pose to the test face. Therefore, the model is learned with the pose information, and a smaller number of selected examples are necessary to represent the test face well. Experimental results in Fig. 10 show that the "IDF + Clustering" training scheme further improves the alignment error, when compared to the non-clustering scheme.

| | Number of Trees | | | | | | | | | |
|---|---|---|---|---|---|---|---|---|---|---|
| | 10 | 20 | 30 | 40 | 50 | 60 | 70 | 80 | 90 | 100 |
| IDF | 680 | 1360 | 2040 | 2720 | 3400 | 4080 | 4760 | 5440 | 6120 | 6800 |
| PCA* | N/A | N/A | N/A | N/A | 768 | 806 | 837 | 852 | 879 | 888 |

Table-3: Feature dimensions of IDF and IDF after using PCA (PCA* means: IDF+PCA, keeping 95% of variance ).

The higher the feature dimension, the more the number of linear-regression weights for the regression model, which requires more computations and more memory space for the prediction step. All the weights of the models for the cascaded stages need to be kept in memory. Another advantage of using IDF, compared to LBF, is that IDF can apply PCA to reduce its feature dimensionality, because IDF is represented by floating-point numbers. From Table 3, we can see that, when the dimension becomes higher, retaining eigenvectors with 95% of variance can reduce the feature dimension by 80% to 90%, while comparable or even better performance can be achieved. Balancing the overhead cost of PCA computation and the relax on linear regression after dimension reduction, theoretically a more optimal and faster solution can be found when the feature dimension of IDF growing up. However, it is hard to implement PCA with LBF feature's binary values. Therefore, IDF with a higher dimensionality can be adopted so as to achieve both efficiency and accuracy, which is hard with LBF feature. Fig. 5 shows the whole proposed algorithm, and the training and fitting stages of our proposed algorithm are described in Algorithm 1 and Algorithm 2, respectively.

**Algorithm 1: IDF Training Stage:**

**Input:** Training data ($I_i$, $S_i$, $\bar{S}_i$) for $i$=1, …, $N$, where $I_i$ represents a face image, $S_i$ is the corresponding shape, and $N$ is the number of training samples.

**Output:** Regressors: $R = (R_1, \dots, R_T)$, $T$: stage count.

**1:** Using $k$-means to cluster shapes in $S$ into $K$ clusters $C = (C_1, \dots, C_K)$, randomly sample initial shapes for each target shape from its belonging cluster $\bar{S}_i \in C_i$ as source shapes;

**2: for** $t$=1 to $T$ **do**

**3:** **for** all $i \in (1 \dots N)$ **do**

**4:** $\Delta S_t^i = S_t^i - \bar{S}_t^i$ ⇒ Calculate shape increment: $\Delta S_t^i$

**5:** $f_t^i = \phi_t(I^i, S_{t-1}^i)$ ⇒ IDF features derived from pose-indexed features

**6:** **end for**

**7:** $R_t = \arg\min_R \sum_i |R(f_t^i) - \Delta S_t^i|$

**8:** **for** all $i \in (1 \dots N)$ **do**

**9:** $\bar{S}_t^i = \bar{S}_t^i + R(f_t^i)$ ⇒ update current shape

**10:** **end for**

**11: end for**

**Algorithm 2: IDF Fitting Stage:**

**Input:** Testing image $I$, initial (mean) shape $S^0$, trained regressors: $R = (R_1, \dots, R_T)$, $T$: stage count.

**Output:** Estimated pose $S^T$

**1: for** $t$=1 to $T$ **do**

**2:** $f_t = \phi_t(I, S_{t-1})$ ⇒ IDF features derived from operation: $\phi_t(I^i, S^i_{t-1})$

**3:** $\Delta S = R_t(\phi_t(I, S_{t-1}))$ ⇒ apply linear regressor $R_t$

**4:** $S_t = S_{t-1} + \Delta S$ ⇒ update pose

**5: end for**

## VI. Experimental Results and Parameter Setting

Using the LBF or IDF feature on random forests can avoid the problem of having insufficient numbers of training samples in the leaf nodes, because these features are not extracted from the samples, but computed from the full binary tree. We have analyzed the encoding process of IDF, and found that the IDF value of each node in the full binary tree is affected by two parameters: the *difference value d* between two brother nodes and the *magnitude value k* for each generation level. However, since the final encoded values of all the nodes are relative values, one of these two parameters can be fixed and another one used for fine-tuning. In our experiments, we fix the value of $d$ to 1, and plot the alignment error curves for different values of $k$.

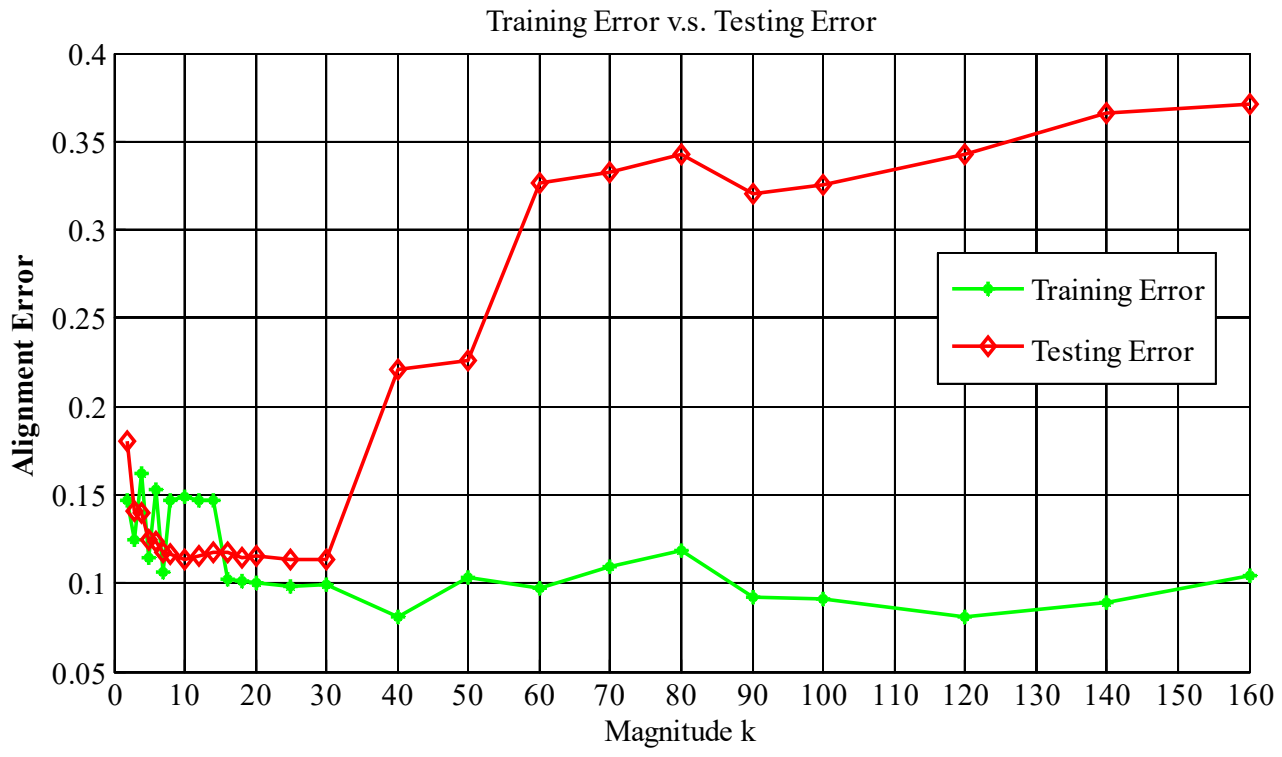

Fig. 12: Alignment errors of different magnitude values of $k$.

As shown in Fig. 12, the alignment errors become the lowest, when the *magnitude value k* is in the range from 10 to 30 (for the tree depth set at 7). This means when the *magnitude value k* is within this range, the encoded values keep the discriminative capability. Therefore, for our proposed IDF feature, the optimal setting is as follows: tree depth: 7, maximum number of stages: 7, number of trees in a forest: 11, number of initialization faces: 50, number of shape clusters: 7, and magnitude value $k$: 10. The trained model, based on our proposed IDF feature and framework, is capable of achieving a comparable alignment quality to state-of-the-art methods [1, 6, 13, 15]. Meanwhile, our algorithm can run at a speed of more than 1,000 frames per second (FPS) on a desktop computer (Intel Core i7 4790 CPU @3.6GHz, 16GB RAM) with C++ code after thread parallelization on 8-core CPUs.

The performance of IDF method, LBF [1], and CLNF [27], in terms of accuracy and the Inter-Occular distance criterion, for different facial landmarks (with 10 facial landmark points) are shown in Fig. 13, which demonstrates that our proposed method IDF is comparable to or outperformances recent state-of-the-art methods. The Fig. 13 also shows that, based on these two criteria, it is very clear all the methods are challenged by the facial landmark points in the mouth area, since the mouth facial landmark points represent significantly changed expression of human faces. For the facial landmark points in the mouth area, our proposed method IDF has clearly improvement than the same regression-based method LBF [1] method and can compare to the classic point distribution model (PDM) based method, the CLNF [27] method.

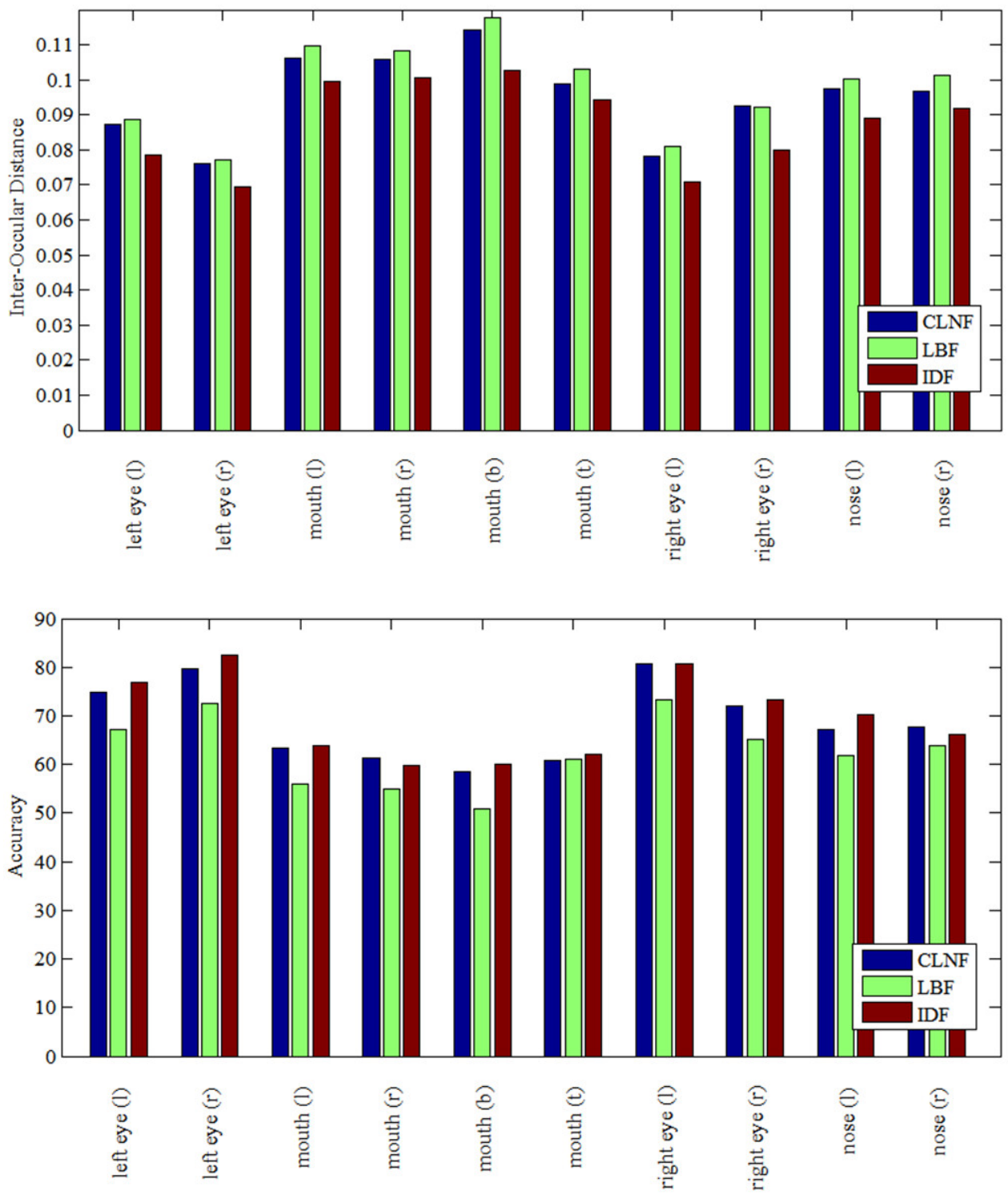

Fig. 13: Comparison of LBF [1], CLNF[27] and IDF, with performance on accuracy and InterOccular distance criterion on 10 facial landmark points in the Helen dataset.

Fig. 14 illustrates some results of the IDF method, and shows that IDF can locate landmarks accurately on faces with different poses and expressions, with occlusion, as well as faces with accessories (glasses). Our proposed method achieves promising performance, compared to the state-of-the-art algorithms [1, 15, 27].

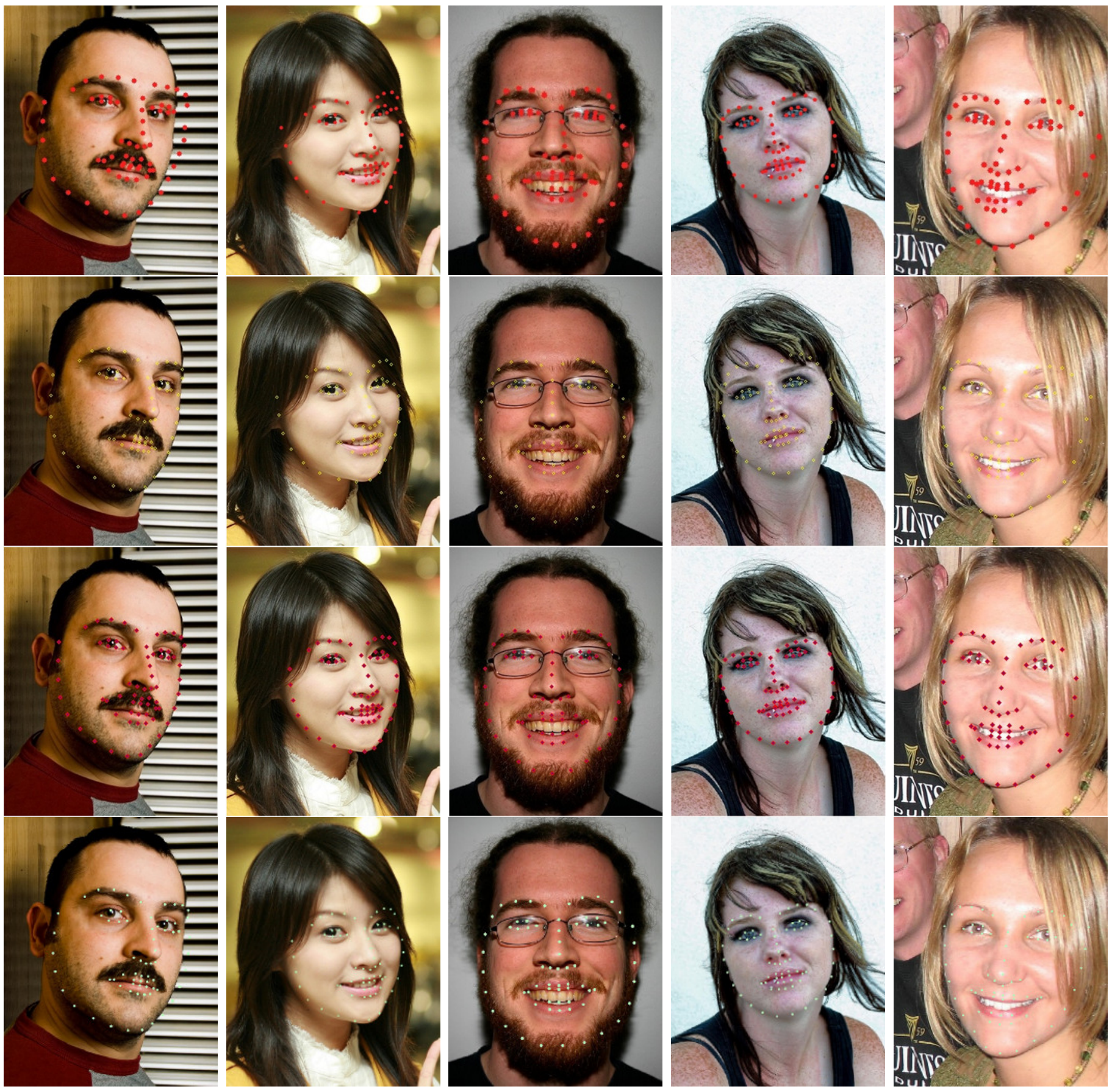

Fig. 14: Fitting results, with 68 landmarks, based on different methods and the Helen dataset:
Row 1: LBF [1], Row 2: One-Milli-Second [15], Row 3: CLNF[27], and Row 4: IDF.

For the linear regression setting, the *LibLinear* package [7] was used for both LBF and IDF, and the linear regression type is set at L2R_L2LOSS_SVR, i.e., $L^2$-regularized $L^2$-loss support vector regression (primal), in which Newton method with trust region step control is employed to achieve faster convergence [30].

## VII. Conclusions and Future Work

Real-time face alignment is still a challenging task. Although many researchers have put efforts into this research area and many algorithms have already been proposed, a highly robust and efficient algorithm is still on the way.

In summary, our proposed method gave contributions on feature extraction of fast face alignment and shape initialization refinement as following: Firstly, in this paper, we have proposed a fast cascaded random-forest regression face-alignment method, based on a novel, simple, but effective and discriminative, feature, which can achieve state-of-the-art performance, in terms of alignment accuracy and computational efficiency. Secondly, we have also addressed the fact that cascaded regression approaches are sensitive to shape initialization. Rather than using a number of blind initializations at the training stage, we propose to initialize shapes, using similar shape samples spanned in different subspaces. With this training strategy, the cascaded regression approach is capable of learning more accurate alignment trajectory and further improving the generalization capability of the trained forests, which result in higher accuracy in the prediction stage. Finally, we have done intensive experimental testing to validate that our proposed method can achieve promising performance both in terms of alignment accuracy and fitting speed. As IDF is a generic random-forest extracted visual feature which can be applied to other computer vision tasks, meanwhile IDF has enriched random-forest based research topics.

IDF is a very efficient feature for face alignment, since it is derived from the efficient pixel-based pose-indexed feature. However, limited by the capacity of pixel-based features, it is more susceptible to image noise, compared to the manually crafted SIFT-based methods. Furthermore, due to the limitation of the random-forest-based cascade regression framework, further work is necessary to tackle these problems for faces with a large pose and occlusion.

## References

[1] Ren, Shaoqing, Xudong Cao, Yichen Wei, and Jian Sun. "Face alignment at 3000 fps via regressing local binary features." In Proceedings of the IEEE Conference on Computer Vision and Pattern Recognition, pp. 1685-1692. 2014.

[2] Zhou, Huiling, Kin-Man Lam, and Xiangjian He. "Shape-appearance-correlated active appearance model." Pattern Recognition 56 (2016): 88-99.

[3] Cootes, Timothy F., and Christopher J. Taylor. "Active shape models—'smart snakes'." In BMVC92, pp. 266-275. Springer London, 1992.

[4] Cootes, Timothy F., Gareth J. Edwards, and Christopher J. Taylor. "Active appearance models." IEEE Transactions on pattern analysis and machine intelligence 23, no. 6 (2001): 681-685.

[5] Dantone, Matthias, Juergen Gall, Gabriele Fanelli, and Luc Van Gool. "Real-time facial feature detection using conditional regression forests." In Computer Vision and Pattern Recognition (CVPR), 2012 IEEE Conference on, pp. 2578-2585. IEEE, 2012.

[6] Xiong, Xuehan, and Fernando De la Torre. "Supervised descent method and its applications to face alignment." In Proceedings of the IEEE conference on computer vision and pattern recognition, pp. 532-539. 2013.

[7] Fan, Rong-En, Kai-Wei Chang, Cho-Jui Hsieh, Xiang-Rui Wang, and Chih-Jen Lin. "LIBLINEAR: A library for large linear classification." Journal of machine learning research 9, no. Aug (2008): 1871-1874.

[8] Dollár, Piotr, Peter Welinder, and Pietro Perona. "Cascaded pose regression." In Computer Vision and Pattern Recognition (CVPR), 2010 IEEE Conference on, pp. 1078-1085. IEEE, 2010.

[9] Chew, Sien W., Patrick Lucey, Simon Lucey, Jason Saragih, Jeffrey F. Cohn, and Sridha Sridharan. "Person-independent facial expression detection using constrained local models." In Automatic Face & Gesture Recognition and Workshops (FG 2011), 2011 IEEE International Conference on, pp. 915-920. IEEE, 2011.

[10] Gao, Hua, Hazım Kemal Ekenel, and Rainer Stiefelhagen. "Pose normalization for local appearance-based face recognition." In International Conference on Biometrics, pp. 32-41. Springer Berlin Heidelberg, 2009.

[11] Wang, Nannan, Dacheng Tao, Xinbo Gao, Xuelong Li, and Jie Li. "A comprehensive survey to face hallucination." International journal of computer vision 106, no. 1 (2014): 9-30.

[12] Viola, Paul, and Michael J. Jones. "Robust real-time face detection." International journal of computer vision 57, no. 2 (2004): 137-154.

[13] Cao, Xudong, Yichen Wei, Fang Wen, and Jian Sun. "Face alignment by explicit shape regression." International Journal of Computer Vision 107, no. 2 (2014): 177-190.

[14] [14] Breiman, Leo. "Random forests." Machine learning 45, no. 1 (2001): 5-32.

[15] Kazemi, Vahid, and Josephine Sullivan. "One millisecond face alignment with an ensemble of regression trees." In Proceedings of the IEEE Conference on Computer Vision and Pattern Recognition, pp. 1867-1874. 2014.

[16] Schulter, Samuel, Paul Wohlhart, Christian Leistner, Amir Saffari, Peter M. Roth, and Horst Bischof. "Alternating decision forests." In Proceedings of the IEEE Conference on Computer Vision and Pattern Recognition, pp. 508-515. 2013.

[17] Moosmann, Frank, Bill Triggs, and Frederic Jurie. "Fast discriminative visual codebooks using randomized clustering forests." In NIPS, vol. 2, p. 4. 2006.

[18] Schulter, Samuel, Christian Leistner, and Horst Bischof. "Fast and accurate image upscaling with super-resolution forests." In Proceedings of the IEEE Conference on Computer Vision and Pattern Recognition, pp. 3791-3799. 2015.

[19] Salvador, Jordi, and Eduardo Pérez-Pellitero. "Naive bayes super-resolution forest." In Proceedings of the IEEE International Conference on Computer Vision, pp. 325-333. 2015.

[20] Belhumeur, Peter N., David W. Jacobs, David J. Kriegman, and Neeraj Kumar. "Localizing parts of faces using a consensus of exemplars." IEEE transactions on pattern analysis and machine intelligence 35, no. 12 (2013): 2930-2940.

[21] Le, Vuong, Jonathan Brandt, Zhe Lin, Lubomir Bourdev, and Thomas S. Huang. "Interactive facial feature localization." In European Conference on Computer Vision, pp. 679-692. Springer Berlin Heidelberg, 2012.

[22] Dalal, Navneet, and Bill Triggs. "Histograms of oriented gradients for human detection." In Computer Vision and Pattern Recognition, 2005. CVPR 2005. IEEE Computer Society Conference on, vol. 1, pp. 886-893. IEEE, 2005.

[23] Ahonen, Timo, Abdenour Hadid, and Matti Pietikainen. "Face description with local binary patterns: Application to face recognition." IEEE transactions on pattern analysis and machine intelligence 28, no. 12 (2006): 2037-2041.

[24] Cristinacce, David, and Timothy F. Cootes. "Feature Detection and Tracking with Constrained Local Models." In BMVC, vol. 1, no. 2, p. 3. 2006.

[25] Cristinacce, David, and Tim Cootes. "Automatic feature localisation with constrained local models." Pattern Recognition 41, no. 10 (2008): 3054-3067.

[26] Matthews, Iain, and Simon Baker. "Active appearance models revisited." International journal of computer vision 60, no. 2 (2004): 135-164.

[27] Baltrusaitis, Tadas, Peter Robinson, and Louis-Philippe Morency. "Constrained local neural fields for robust facial landmark detection in the wild." In Proceedings of the IEEE International Conference on Computer Vision Workshops, pp. 354-361. 2013.

[28] Saragih, Jason M., Simon Lucey, and Jeffrey F. Cohn. "Deformable model fitting by regularized landmark mean-shift." International Journal of Computer Vision 91, no. 2 (2011): 200-215.

[29] Baker, Simon, and Iain Matthews. "Lucas-kanade 20 years on: A unifying framework." International journal of computer vision 56, no. 3 (2004): 221-255.

[30] Lin, Chih-Jen, Ruby C. Weng, and S. Sathiya Keerthi. "Trust region newton method for logistic regression." Journal of Machine Learning Research 9, no. Apr (2008): 627-650.

[31] Asthana, Akshay, Stefanos Zafeiriou, Shiyang Cheng, and Maja Pantic. "Robust discriminative response map fitting with constrained local models." In Proceedings of the IEEE Conference on Computer Vision and Pattern Recognition, pp. 3444-3451. 2013.

[32] Zhou, Feng, Jonathan Brandt, and Zhe Lin. "Exemplar-based graph matching for robust facial landmark localization." In Proceedings of the IEEE International Conference on Computer Vision, pp. 1025-1032. 2013.

[33] Uijlings, Jasper RR, Koen EA Van De Sande, Theo Gevers, and Arnold WM Smeulders. "Selective search for object recognition." International journal of computer vision 104, no. 2 (2013): 154-171.

[34] Liu, Wei, Dragomir Anguelov, Dumitru Erhan, Christian Szegedy, Scott Reed, Cheng-Yang Fu, and Alexander C. Berg. "SSD: Single shot multibox detector." In European Conference on Computer Vision, pp. 21-37. Springer International Publishing, 2016.

[35] Redmon, Joseph, Santosh Divvala, Ross Girshick, and Ali Farhadi. "You only look once: Unified, real-time object detection." In Proceedings of the IEEE Conference on Computer Vision and Pattern Recognition, pp. 779-788. 2016.

[36] Weng, Renliang, Jiwen Lu, Yap-Peng Tan, and Jie Zhou. "Learning Cascaded Deep Auto-Encoder Networks for Face Alignment." IEEE Transactions on Multimedia 18, no. 10 (2016): 2066-2078.

[37] Zhang, Jie, Shiguang Shan, Meina Kan, and Xilin Chen. "Coarse-to-fine auto-encoder networks (cfan) for real-time face alignment." In European Conference on Computer Vision, pp. 1-16. Springer International Publishing, 2014.

[38] Cireşan, Dan Claudiu, Ueli Meier, Luca Maria Gambardella, and Jürgen Schmidhuber. "Deep, big, simple neural nets for handwritten digit recognition." Neural computation 22, no. 12 (2010): 3207-3220.

[39] Wang, Anran, Jiwen Lu, Jianfei Cai, Tat-Jen Cham, and Gang Wang. "Large-margin multi-modal deep learning for RGB-D object recognition." IEEE Transactions on Multimedia 17, no. 11 (2015): 1887-1898.

[40] Zhang, Zhanpeng, Ping Luo, Chen Change Loy, and Xiaoou Tang. "Facial landmark detection by deep multi-task learning." In European Conference on Computer Vision, pp. 94-108. Springer International Publishing, 2014.

[41] Dong, Chao, Chen Change Loy, Kaiming He, and Xiaoou Tang. "Learning a deep convolutional network for image super-resolution." In European Conference on Computer Vision, pp. 184-199. Springer International Publishing, 2014.

[42] Kim, Jiwon, Jung Kwon Lee, and Kyoung Mu Lee. "Accurate image super-resolution using very deep convolutional networks." In Proceedings of the IEEE Conference on Computer Vision and Pattern Recognition, pp. 1646-1654. 2016.

[43] Ledig, Christian, Lucas Theis, Ferenc Huszár, Jose Caballero, Andrew Cunningham, Alejandro Acosta, Andrew Aitken et al. "Photo-realistic single image super-resolution using a generative adversarial network." arXiv preprint arXiv:1609.04802 (2016)..

[44] Wang, Naiyan, and Dit-Yan Yeung. "Learning a deep compact image representation for visual tracking." In Advances in neural information processing systems, pp. 809-817. 2013.

[45] Abdulnabi, Abrar H., Gang Wang, Jiwen Lu, and Kui Jia. "Multi-task CNN model for attribute prediction." IEEE Transactions on Multimedia 17, no. 11 (2015): 1949-1959.